\documentclass[letterpaper]{article} 
\usepackage{aaai24}  
\usepackage{times}  
\usepackage{helvet}  
\usepackage{courier}  
\usepackage[hyphens]{url}  
\usepackage{graphicx} 
\usepackage{natbib}  
\usepackage{caption} 
\usepackage{algorithm}
\usepackage{algorithmic}
\usepackage{amsmath}
\usepackage{multirow}
\usepackage{bbding}
\usepackage{pifont}
\usepackage{wasysym}
\usepackage{amssymb}
\usepackage{booktabs}

\usepackage{newfloat}
\usepackage{listings}
\DeclareCaptionStyle{ruled}{labelfont=normalfont,labelsep=colon,strut=off} 
\floatstyle{ruled}
\newfloat{listing}{tb}{lst}{}
\floatname{listing}{Listing}
\title{Modeling Stereo-Confidence Out of the End-to-End Stereo-Matching Network \\via Disparity Plane Sweep}
\author{
    Jae Young Lee\textsuperscript{\rm 1}\equalcontrib,
    Woonghyun Ka\textsuperscript{\rm 2}\equalcontrib,
    Jaehyun Choi\textsuperscript{\rm 1},
    Junmo Kim\textsuperscript{\rm 1}
}
\affiliations{
    \textsuperscript{\rm 1}School of Electrical Engineering, KAIST, Daejeon, South Korea\\
    \textsuperscript{\rm 2}Hyundai Motor Company, Seoul, South Korea\\
    \{mcneato, chlwogus, junmo.kim\}@kaist.ac.kr, kwh950724@hyundai.com
}

\usepackage{bibentry}

\begin{document}

\maketitle

\begin{abstract}
We propose a novel stereo-confidence that can be measured externally to various stereo-matching networks, offering an alternative input modality choice of the cost volume for learning-based approaches, especially in safety-critical systems.
Grounded in the foundational concepts of disparity definition and the disparity plane sweep, the proposed stereo-confidence method is built upon the idea that any shift in a stereo-image pair should be updated in a corresponding amount shift in the disparity map. 
Based on this idea, the proposed stereo-confidence method can be summarized in three folds.
1) Using the disparity plane sweep, multiple disparity maps can be obtained and treated as a 3-D volume (predicted disparity volume), like the cost volume is constructed. 
2) One of these disparity maps serves as an anchor, allowing us to define a desirable (or ideal) disparity profile at every spatial point.
3) By comparing the desirable and predicted disparity profiles, we can quantify the level of matching ambiguity between left and right images for confidence measurement. 
Extensive experimental results using various stereo-matching networks and datasets demonstrate that the proposed stereo-confidence method not only shows competitive performance on its own but also consistent performance improvements when it is used as an input modality for learning-based stereo-confidence methods.
\end{abstract}

\section{Introduction}

Stereo-matching is a fundamental task in computer vision that aims to estimate the depth information of a scene from a pair of stereo images. 
Accurate depth estimation is essential for a wide range of applications, such as autonomous driving, robotics, and augmented reality.
Over decades, various stereo-matching methods have been proposed, ranging from traditional methods based on hand-crafted features to more recent learning-based methods.
These methods typically generate the cost volume or feature maps to estimate the disparity value for each pixel. 
Despite impressive results achieved by recent learning-based stereo-matching methods, they occasionally fail in ill-posed regions such as occlusion boundaries, repeated patterns, textureless regions, and non-Lambertian surfaces due to inherent ambiguities tied to the correspondence problem~\cite{StereoMatchingSurvey}.

To deal with such ambiguous correspondences, stereo-confidence measurement has been utilized to identify whether the results of stereo-matching methods are reliable at each pixel.
In the past, conventional approaches commonly measure confidence by analyzing the cost volume obtained from stereo-matching methods.
With the rise of the recent deep learning paradigm, learning-based stereo-confidence approaches have also been widely studied leveraging various combinations of input modalities~\cite{ConfidenceSurvey}. 
Especially, in the case of LAF-Net~\cite{LAF-Net}, it was the first one among learning-based approaches to leverage the cost volume as an input modality, achieving state-of-the-art performance through tri-modal inputs.
Both conventional and learning-based approaches have demonstrated that the cost volume is an effective input cue among input modalities.
The cost profile in the cost volume provides information about how confident the stereo-matching method is in its predictions by matching cost (or probability) for each disparity plane shift.
The cost volume can be further utilized for tasks such as uncertainty estimations~\cite{UncertaintyStereo} and refinement of stereo-matching networks~\cite{ConfPropagation}. 

However, the cost volumes (or internal features) are being encapsulated within the stereo-matching network itself due to the recent trend of stereo-matching networks operating in an end-to-end manner.
Furthermore, especially in safety-critical systems closely tied to the stereo-matching networks, the architecture is kept confidential, and access to its internal information is restricted to protect the network from potential external threats~\cite{SafetyCritical,StereoAdversarial1,StereoAdversarial2}.
As a result, this prompts the necessity of the exploration of estimating confidence without accessing the cost volumes (or internal features) and from outside the stereo-matching network~\cite{selfadapting}.

Unfortunately, the cost volume is an effective cue in stereo-confidence estimation.
Thus, in this paper, we aim to quantify the level of matching ambiguity from the external stereo-matching method and achieve effects comparable to using the cost volume as an input without performance degradation.
Furthermore, we intend to introduce the proposed confidence as an alternative input of the cost volume to learning-based stereo-confidence methods, making it suitable for use even in safety-critical systems.
It can be achieved by reinterpreting confidence based on the relationship between the cost volume and the multiple disparity maps.
The proposed method is built upon two main concepts: the definition of disparity and the disparity plane sweep.
In stereo-matching, the disparity is defined as a displacement between a point in the left image and its corresponding point in the right image of a stereo-image pair.
The disparity plane sweep consecutively shifts an image with respect to the reference image in stereo images, which is generally used to construct the cost volume to find the correspondence between a stereo-image pair in stereo-matching. 
By the definition of disparity itself, any shift by the disparity plane sweep in a stereo-image pair indicates that the disparity map should be updated in line with the corresponding shift amount.
Based on the definition of the disparity and the disparity plane sweep, the proposed stereo-confidence method can be condensed into three main aspects:
1) just as the cost volume is constructed, multiple disparity maps can be obtained from any stereo-matching network and treated as a 3-D volume (disparity volume) using the disparity plane sweep. 
2) Using an obtained disparity map without shift as an anchor, the desirable disparity profile can be defined and treated as an ideal one, as the ideal cost profile is defined.
3) By comparing the desirable and predicted disparity profiles at every spatial point in the disparity volume, we can quantify the level of matching ambiguity between left and right images for confidence measurement. 

To demonstrate the effectiveness of the proposed method not only as confidence but also as an input modality, we compare the performance of the proposed confidence to the existing learning-based stereo-confidence methods. 
Furthermore, we present the experimental results when the proposed method is leveraged as an additional input modality for the existing learning-based stereo-confidence methods.
While the reinterpreted confidence might seem simple, and its measurement is a conventional method, the experimental results consistently demonstrate the effectiveness of the proposed method across diverse datasets, regardless of the type of the stereo-matching networks. 
Furthermore, the proposed method shows potential in safety-critical systems as it successfully works as an alternative choice of the cost volume.

\section{Related Works}
In this section, the conventional and learning-based stereo-confidence methods are briefly introduced.

\subsection{Conventional Methods}
The conventional methods have been extensively studied over the past few decades and mainly have relied on analyzing the matching cost volume. 
\citet{stereo_conf} thoroughly investigated 17 confidence measures, focusing on conventional methods. 
These methods developed algorithms based on the minimum cost and local properties of the cost profile~\cite{conventional_local0,conventional_local1,conventional_local2,conventional_local3,conventional_local4,conventional_local5}. 
Some conventional methods examine the entire curve of the cost profile to extract useful information for measuring confidence~\cite{conventional_local1,conventional_curve0,conventional_curve1,conventional_curve2}.
Unlike these methods that concentrate on the cost profile inside the stereo-matching model, the proposed approach examines the disparity profile outside the stereo-matching network, making it suitable for learning-based end-to-end stereo-matching methods.

\subsection{Learning-Based Methods}
Recently, learning-based methods have used various deep-learning techniques to measure confidence.
These methods extract features directly from input modalities, such as the disparity maps, reference image, and cost volume, and then estimate the confidence from features. 
Various learning-based methods have been proposed using different combinations of input modalities.
\textbf{Single-modality (disparity)}, CCNN~\cite{CCNN} measured confidence using a convolutional neural network for the first time.
\citet{meta-confidence} proposed meta-confidence to improve stereo-confidence quality by encoding the reliability of confidence once more.
\textbf{Bi-modality (disparity, image)}, LFN~\cite{LFN} pioneered the stereo-confidence with the bi-modal inputs and fusion strategies.
\citet{LGC-Net} proposed ConfNet to obtain global confidence with large receptive fields.
They also proposed LGC-Net, a local-global confidence framework, combining ConfNet with local confidence network, such as CCNN and LFN. 
\textbf{Tri-modality (disparity, image, cost volume)}, LAF-Net~\cite{LAF-Net} achieved remarkable performance with tri-modal inputs fused by an attention mechanism.

Nevertheless, a notable limitation of these methods is their reliance on ground truth disparity maps during the training phase.
As a result, their performance may be sub-optimal when faced with out-of-distribution data domains. 
In contrast, our proposed approach itself, like a conventional method, is independent of training processes.


\begin{figure}[!t]
    \centering
    \includegraphics[width=0.92\columnwidth]{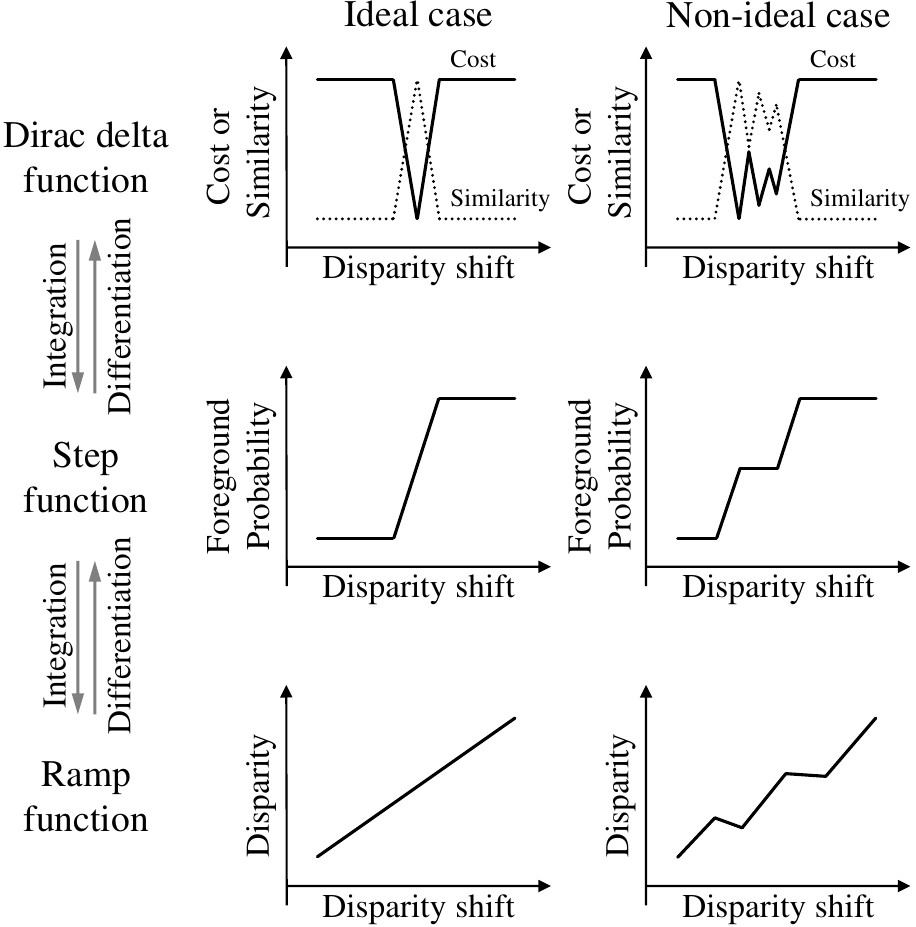}
    \caption{Conceptual description of ideal and non-ideal profiles according to the disparity plane sweep in signal processing perspective.}
    \label{fig:reinterpret}
\end{figure}

\section{Method}
\begin{figure*}[!t]
    \centering
    \includegraphics[width=\linewidth]{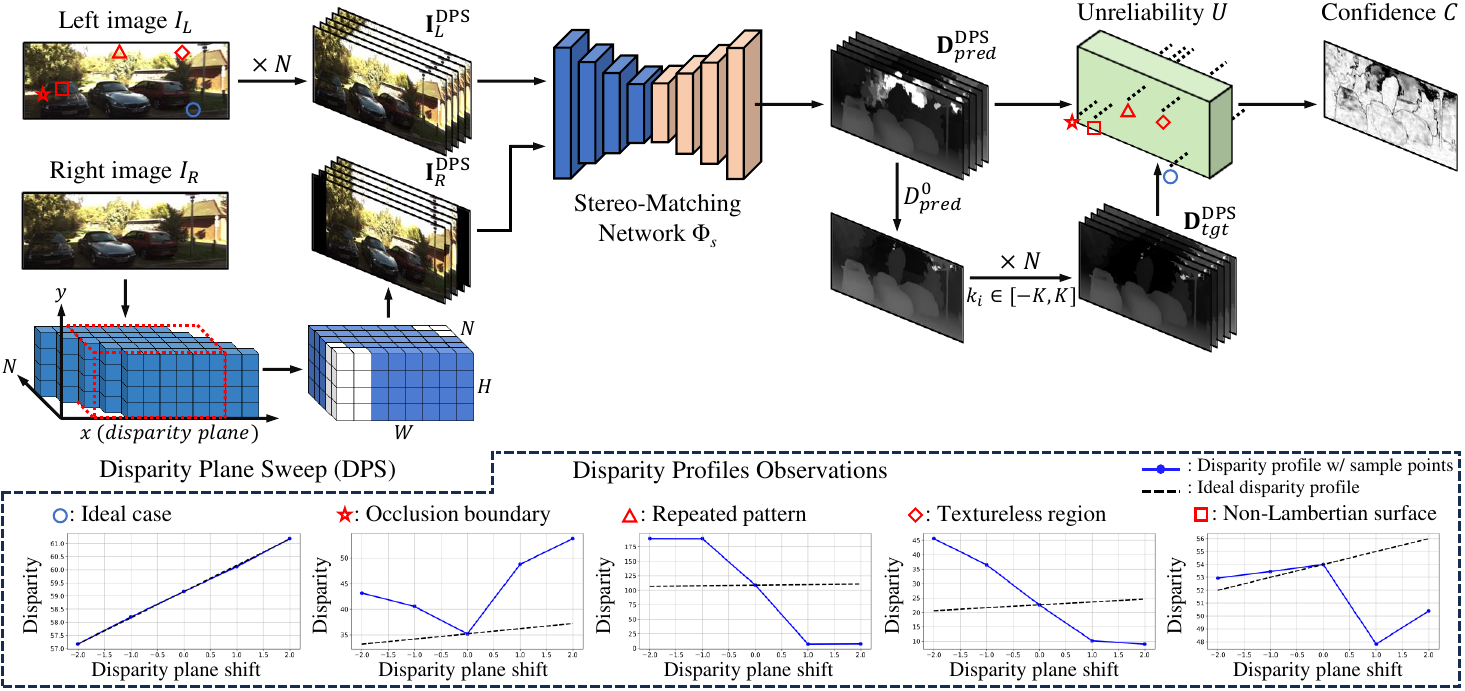}
    \caption{The entire process of the proposed method for quantifying unreliability (i.e., matching ambiguity) and measuring confidence out of the stereo-matching network using disparity plane sweep and the observations of disparity profiles sampled from corresponding pixels in the left image $I_{L}$.}
    \label{fig:architecture}
\end{figure*}
In this section, first, we revisit the definition of disparity and briefly explain the concept of the proposed method. 
Second, to reinterpret confidence, a property of the disparity profile derived from the relationship between the cost volume and the disparity volume is presented. 
Third, the process to obtain the disparity volume is explained.
Then, the proposed stereo-confidence method is described.

\subsection{Preliminaries and Concept}
In stereo imaging, points from 3D real-world coordinates are projected onto different pixel coordinates within the left and right images of a stereo-image pair.
In terms of pixel units, the disparity refers to the horizontal displacement between a pixel in the left image and a corresponding pixel in the right image when overlaying the stereo images.
Using the left image as the reference, the stereo-matching aims to estimate the disparity values within a given stereo-image pair.
Both conventional and learning-based approaches rely on visual similarity (or difference) to find image regions identical to the reference image regions, which are determined by the cost obtained by comparing the pixel intensities or features of image patches.
Through disparity plane sweep, i.e., shifting the right image concerning the reference image within a predefined maximum disparity range, the costs between the reference image and the shifted right image are calculated. 
The accumulated costs result in the form of a 3D cost volume.
Then, the disparity plane shift with the lowest cost is determined as the disparity value.
However, as the cost is computed within an image patch (or limited receptive field), stereo-matching frequently encounters challenges in regions where the correspondence is ambiguous, such as occlusion boundaries, repeated patterns, textureless regions, and non-Lambertian surfaces \cite{StereoMatchingSurvey}.
In essence, ambiguous correspondences in the stereo-matching result in the misidentification of image patches, causing unreliable disparity estimations. 
Conversely, clear correspondences yield trustworthy results.

To quantify matching ambiguity, the proposed method utilizes two fundamental components: a definition of the disparity and the disparity plane sweep.
The outline of the proposed method can be summarized in three folds.
1) Like the cost volume is constructed, multiple disparity maps in the form of 3-D volume can be obtained using the disparity plane sweep.
2) Using a disparity map obtained by zero shift as an anchor, the desirable disparity profile in 3-D disparity volume can be defined and treated as an ideal one, as an ideal cost profile is defined.
3) Matching ambiguity can be quantified by comparing the desirable and obtained disparity profiles at every spatial location in the disparity volume.

\subsection{Reinterpretation of Confidence}
The relationship between the cost volume and disparity maps is introduced in~\citet{LFUDM}.
In the light field (LF), they \cite{FBS-SFA, LFUDM} showed the relationship among the cost volume-based \cite{SPO,SPO-MO,CAE,LF_OCC}, foreground-background separation (FBS)-based \cite{FBS-AO,FBS-JSTSP,FBSOCC-SFA}, and depth model-based methods \cite{FocalStackNet,fusionnet,epinet} from a signal processing perspective (See Fig.~\ref{fig:reinterpret}) according to the LF parameterization, which corresponds to the disparity plane sweep in the stereo-matching.
Similar to the LF, stereo can have such a relationship among the cost volume-based \cite{MC-CNN}, FBS-based \cite{badki2020Bi3D}, and depth model-based methods \cite{PSMNet,GANet,AANet} according to the disparity plane sweep.
Notably, the majority of recent stereo-matching networks operate in an end-to-end manner, with such end-to-end networks predominantly falling into the group of depth model-based methods.

In the stereo-matching methods, the disparity plane sweep is generally utilized to obtain the cost volume, which is an internal feature in their networks. 
Differently, in this paper, using the disparity plane sweep, multiple disparity maps from the end-to-end stereo-matching networks are obtained to measure the confidence of disparity maps.
Recall that the disparity is defined as the pixel-unit horizontal displacement between corresponding pixels when overlaying the stereo-image pair.
According to the definition of disparity, when the right image is horizontally shifted by $k$ pixels with respect to the left (reference) image, the disparity map obtained by an end-to-end stereo-matching model $\Phi_{s}$ should change by $k$ pixels under the identical correspondence as follows:
\begin{equation}
    \Phi_{s}(I_{L}, I_{R}) + k = \Phi_{s}(I_{L}, I_{R}^{k}),
\end{equation}
where $I_{L}$, $I_{R}$, and $I_{R}^{k}$ denote the left image, the right image, and the right image horizontally shifted by $k$, respectively. 

Ideally, according to the disparity plane sweep, the disparity profile should be shaped into a linear line, resembling a ramp function as shown in Fig.~\ref{fig:reinterpret}. 
However, in real-world scenarios, if the right image is shifted using the disparity plane sweep, stereo-matching methods often fail to maintain identical correspondence due to ambiguous correspondence.
The ambiguous correspondence results in the distortion of the disparity profile.
At the bottom of Fig.~\ref{fig:architecture}, examples of disparity profiles are presented (See supplementary material for more examples). 
While the disparity profile of the ideal case is shaped into a linear line, that of the occlusion boundary, repeated pattern, textureless region, and non-Lambertian surface is distorted. 
Based on the observations of the disparity profile, we reinterpret the conventional stereo-confidence methods. 
While the conventional stereo-confidence methods analyze the cost profile with respect to the ideal cost profile (Dirac-delta), the proposed method analyzes the disparity profile with respect to the ideal disparity profile, i.e., linear line (ramp).
Using the disparity map obtained without shift as an anchor, we measure whether the disparity profile is shaped into a linear line or not. 
With respect to the linear line anchored by the zero-shifted disparity map, a low distortion indicates reduced ambiguity and high confidence. 
Conversely, a high distortion signifies increased ambiguity and low confidence.

\begin{table}[t]
\centering
\resizebox{0.7\columnwidth}{!}
{
\begin{tabular}{crrr}
\hline
\multirow{2}{*}{Method}			            & \multicolumn{3}{c}{Input Modality Type}          								                                    \\ \cline{2-4}
								            & \multicolumn{1}{c}{Disparity}		& \multicolumn{1}{c}{Image}	& \multicolumn{1}{c}{Cost Volume}			\\ \hline
\multicolumn{1}{l}{Ours}	            & \multicolumn{1}{c}{Muliple}				& 								 	&									\\ 
\multicolumn{1}{l}{CCNN}		    & \multicolumn{1}{c}{Single}				& 									&									\\ 
\multicolumn{1}{l}{CCNN$^{\dagger}$}		    & \multicolumn{1}{c}{Muliple}				& 									&									\\ \hline
\multicolumn{1}{l}{LFN}			    & \multicolumn{1}{c}{Single}				& \multicolumn{1}{c}{\checkmark}	&									\\ 
\multicolumn{1}{l}{ConfNet}		    & \multicolumn{1}{c}{Single}				& \multicolumn{1}{c}{\checkmark}	&									\\ 
\multicolumn{1}{l}{LGC-Net}		    & \multicolumn{1}{c}{Single}				& \multicolumn{1}{c}{\checkmark} 	&									\\ 
\multicolumn{1}{l}{LAF-Net$^{*}$}		    & \multicolumn{1}{c}{Single}				& \multicolumn{1}{c}{\checkmark} 	&									\\ 
\multicolumn{1}{l}{LAF-Net$^{*\dagger}$}		            & \multicolumn{1}{c}{Multiple}			& \multicolumn{1}{c}{\checkmark} 	&									\\ \hline
\multicolumn{1}{l}{LAF-Net}		    & \multicolumn{1}{c}{Single}				& \multicolumn{1}{c}{\checkmark} 	& \multicolumn{1}{c}{\checkmark} 	\\ 
\multicolumn{1}{l}{LAF-Net$^{\dagger}$}		            & \multicolumn{1}{c}{Multiple}			& \multicolumn{1}{c}{\checkmark} 	& \multicolumn{1}{c}{\checkmark} 	\\ \hline
\end{tabular}
}
\caption{Various combinations of input modalities according to the stereo-confidence methods in our experiments. The LAF-Net$^*$ denotes LAF-Net that uses bi-modal inputs (disparity, image) without cost volume. The superscript ${\dagger}$ denotes the methods using the proposed method as an additional input modality.}
\label{tab:input_mod}
\end{table}

\subsection{Obtaining Disparity Profiles}
Fig.~\ref{fig:architecture} describes an entire pipeline of the proposed method to obtain the stereo-confidence.
First, with input stereo-image pair $I_{L},I_{R}\in \mathbb{R}^{3\times H\times W}$ and the number of disparity plane shifts $N$, we generate a set of disparity plane swept right images $\textbf{I}_{R}^{\text{DPS}}=\{I_{R}^{k_{i}}\in \mathbb{R}^{3\times H\times W}\mid i=1,2,...,N \}$ by concatenating $N$ right images shifted by $k_{i}$ pixels obtained from the disparity plane sweep in range of $k_{i}\in[-K,K]$.
A set of reference images, $\textbf{I}_{L}^{\text{DPS}}$ is simply generated by repeating the left image $I_{L}$ by $N$ times.
Then, $\textbf{I}_{L}^{\text{DPS}}$ and $\textbf{I}_{R}^{\text{DPS}}\in \mathbb{R}^{N\times 3\times H\times W}$ have the same dimensions, and they go through a pre-trained end-to-end stereo-matching network $\Phi_{s}$.
By doing so, a set of predicted disparity maps $\textbf{D}_{pred}^{\text{DPS}}$ can be obtained in the form of the disparity volume as follows:
\begin{equation}
    \textbf{D}_{pred}^{\text{DPS}}=\{D_{pred}^{k_{i}}\in \mathbb{R}^{H\times W}\mid i=1,...,N\}=\Phi_{s}(\textbf{I}_{L}^{\text{DPS}}, \textbf{I}_{R}^{\text{DPS}}).
\end{equation}
Based on a zero-shifted disparity map $D_{pred}^0$, a set of target disparity maps $\mathbf{D}_{tgt}^{\text{DPS}}$ is generated by adding $k_i$ in the form of the disparity volume as follows:
\begin{equation}
    \textbf{D}_{tgt}^{\text{DPS}}=\{D_{pred}^{0}+k_{i}\in \mathbb{R}^{H\times W}\mid i=1,2,...,N\}. 
\end{equation}
For the main experiments, $N$ and $K$ are set to 5 and 2, respectively.

\subsection{Proposed Stereo-Confidence Method}
Using the sets of predicted and target disparity maps $\textbf{D}_{pred}^{\text{DPS}},\textbf{D}_{tgt}^{\text{DPS}}\in \mathbb{R}^{N\times H\times W}$, we can measure the degree of distortion, which represents the level of ambiguity, the unreliability $U(p)\in \mathbb{R}^{1\times H\times W}$ at each pixel $p$ as follows:
\begin{equation}
    U(p) = \frac{1}{N-1}\sum\lVert \textbf{D}_{tgt}^{\text{DPS}}(p) - \textbf{D}_{pred}^{\text{DPS}}(p) \rVert_{1}.
\end{equation}
Using the unreliability $U(p)$, the confidence $C(p)$ at each pixel $p$ can be obtained by
\begin{equation} 
    C(p)=e^{-\sigma \frac{U(p)}{d_{max}}}, 
\end{equation}
where a scale factor $\sigma$ is set to have a confidence of 0.5 when the unreliability $U(p)$ is 1 and $d_{max}$ is the maximum disparity value of $\Phi_{s}$, respectively.

\section{Experiments}

\begin{table*}[t]
\centering
\resizebox{\linewidth}{!}
{%
\begin{tabular}{clrrrrrrrrrrr}
\hline
\multicolumn{1}{c}{\multirow{2}{*}[-0.7ex]{Dataset}} & \multicolumn{1}{c}{\multirow{2}{*}[-0.7ex]{Stereo}} & \multicolumn{3}{c}{\multirow{1}{*}[-0.4ex]{Single-Modality}} & \multicolumn{5}{c}{\multirow{1}{*}[-0.4ex]{Bi-Modality}} & \multicolumn{2}{c}{\multirow{1}{*}[-0.4ex]{Tri-Modality}} & \multicolumn{1}{c}{\multirow{2}{*}[-0.7ex]{Optimal}} \\\cmidrule(rl){3-5}\cmidrule(rl){6-10}\cmidrule(rl){11-12}
                          & \multicolumn{1}{c}{} & \multicolumn{1}{c}{\multirow{1}{*}[+0.15ex]{Ours}} & \multicolumn{1}{c}{\multirow{1}{*}[+0.15ex]{CCNN}} & \multicolumn{1}{c}{\multirow{1}{*}[+0.3ex]{CCNN$^{\dagger}$}} & \multicolumn{1}{c}{\multirow{1}{*}[+0.15ex]{LFN}} & \multicolumn{1}{c}{\multirow{1}{*}[+0.15ex]{ConfNet}} & \multicolumn{1}{c}{\multirow{1}{*}[+0.15ex]{LGC-Net}} & \multicolumn{1}{c}{\multirow{1}{*}[+0.15ex]{LAF-Net$^{*}$}} & \multicolumn{1}{c}{\multirow{1}{*}[+0.3ex]{LAF-Net$^{*\dagger}$}} & \multicolumn{1}{c}{\multirow{1}{*}[+0.15ex]{LAF-Net}} & \multicolumn{1}{c}{\multirow{1}{*}[+0.3ex]{LAF-Net$^{\dagger}$}} & \multicolumn{1}{c}{}                        \\\hline
\multirow{7}{*}{K2012}    & PSMNet                                                     & \underline{0.1659}                   & 0.2748                   & \textbf{0.1306}                   & 0.4182                  & 0.8125                      & 0.2751                      & \underline{0.2138}                      & \textbf{0.1084}                     & \underline{0.0982}                      & \textbf{0.0903}                      & 0.0114                                        \\
                          & GANet                                                      & \underline{0.1094}                   & 0.1411                   & \textbf{0.0737}                   & 0.1514                  & 0.5348                      & 0.1615                      & \underline{0.1225}                      & \textbf{0.0632}                     & \underline{0.0540}                      & \textbf{0.0522}                      & 0.0052                                        \\
                          & STTR                                                       & 0.2036                   & \underline{0.1810}                   & \textbf{0.1467}                   & 0.1766                  & 0.6639                      & 0.1943                      & \underline{0.1653}                      & \textbf{0.1285}                     & \underline{0.1182}                      & \textbf{0.1166}                      & 0.0558                                        \\
                          & RAFT                                                       & \underline{0.4729}                   & 0.6997                   & \textbf{0.4129}                   & 1.0849                  & 1.3327                      & 0.8685                      & \underline{0.7820}                      & \textbf{0.3563}                     & \underline{0.3862}                      & \textbf{0.3414}                      & 0.0776                                        \\
                          & LEAStereo                                                  & \underline{0.1886}                   & 0.2636                   & \textbf{0.1569}                   & 0.3217                  & 0.8333                      & 0.2506                      & \underline{0.2469}                      & \textbf{0.1284}                     & \underline{0.1059}                      & \textbf{0.1030}                      & 0.0146                                        \\
                          & ACVNet                                                     & \underline{0.1189}                   & 0.1848                   & \textbf{0.0915}                   & 0.3426                  & 0.7476                      & 0.2959                      & \underline{0.1862}                      & \textbf{0.0747}                     & \underline{0.0798}                      & \textbf{0.0726}                      & 0.0107                                        \\
                          & IGEV                                                       & \underline{0.0856}                   & 0.1327                   & \textbf{0.0592}                   & 0.1770                  & 0.4834                      & 0.1544                      & \underline{0.1462}                      & \textbf{0.0552}                     & \underline{0.0819}                      & \textbf{0.0555}                      & 0.0057                                        \\ \hline
\multirow{7}{*}{K2015}    & PSMNet                                                     & \underline{1.2924}                   & 1.5767                   & \textbf{1.1146}                   & 1.7879                  & 2.6216                      & 1.5192                      & \underline{1.3849}                      & \textbf{1.1574}                     & \underline{1.1475}                      & \textbf{1.0751}                      & 0.4152                                        \\
                          & GANet                                                      & \underline{0.2639}                   & 0.3641                   & \textbf{0.1798}                   & 0.3439                  & 0.8074                      & 0.4174                      & \underline{0.3126}                      & \textbf{0.1789}                     & \underline{0.1446}                      & \textbf{0.1429}                      & 0.0231                                        \\
                          & STTR                                                       & 0.2027                   & \underline{0.2006}                   & \textbf{0.1531}                   & 0.2040                  & 0.5377                      & 0.2314                      & \underline{0.1719}                      & \textbf{0.1308}                     & \underline{0.1215}                      & \textbf{0.1196}                      & 0.0454                                        \\
                          & RAFT                                                       & \underline{0.3692}                   & 0.4925                   & \textbf{0.3110}                   & 0.8226                  & 0.7721                      & 0.6607                      & \underline{0.4071}                      & \textbf{0.2632}                     & \underline{0.3087}                      & \textbf{0.2507}                      & 0.0409                                        \\
                          & LEAStereo                                                  & \underline{0.4809}                   & 0.5735                   & \textbf{0.3955}                   & 0.7827                  & 1.4229                      & 0.6322                      & \underline{0.5308}                      & \textbf{0.3954}                     & \underline{0.3875}                      & \textbf{0.3448}                      & 0.1090                                        \\
                          & ACVNet                                                     & \underline{0.7263}                   & 0.9087                   & \textbf{0.6202}                   & 1.1663                  & 1.6532                      & 1.1239                      & \underline{0.7437}                      & \textbf{0.6119}                     & \underline{0.5824}                      & \textbf{0.5572}                      & 0.2351                                        \\
                          & IGEV                                                       & \underline{0.0752}                   & 0.1404                   & \textbf{0.0566}                   & 0.1239                  & 0.2857                      & 0.1214                      & \underline{0.1080}                      & \textbf{0.0504}                     & \underline{0.0700}                      & \textbf{0.0555}                      & 0.0039                                        \\ \hline
\multirow{7}{*}{\shortstack[c]{VK2-S6}}  & PSMNet                                                     & \underline{0.9487}                   & 1.7111                   & \textbf{0.8283}                   & 2.0130                  & 3.8305                      & 1.4161                      & \underline{0.9103}                      & \textbf{0.5897}                     & \underline{0.6592}                      & \textbf{0.6378}                      & 0.1688                                        \\
                          & GANet                                                      & 0.7156                   & \underline{0.5764}                   & \textbf{0.4316}                   & 0.5689                  & 2.1518                      & 0.4978                      & \underline{0.4343}                      & \textbf{0.2716}                     & \underline{0.3757}                      & \textbf{0.3393}                      & 0.0504                                        \\
                          & STTR                                                       & \underline{0.6934}                   & 0.7067                   & \textbf{0.5851}                   & 1.2171                  & 2.9325                      & \underline{0.7810}                      & 0.9654                      & \textbf{0.4719}                     & \underline{0.4864}                      & \textbf{0.4646}                      & 0.1683                                        \\
                          & RAFT                                                       & \underline{0.4470}                   & 0.6569                   & \textbf{0.3617}                   & 1.0762                  & 1.1631                      & 0.6991                      & \underline{0.4815}                      & \textbf{0.2636}                     & \underline{0.4420}                      & \textbf{0.3108}                      & 0.0636                                        \\
                          & LEAStereo                                                  & 0.4804                   & \underline{0.4590}                   & \textbf{0.3765}                   & 1.2508                  & 2.5696                      & 0.6148                      & \underline{0.3155}                      & \textbf{0.2615}                     & \underline{0.3755}                      & \textbf{0.2915}                      & 0.0550                                        \\
                          & ACVNet                                                     & \underline{0.6345}                   & 1.1484                   & \textbf{0.4208}                   & 3.1008                  & 2.6430                      & 2.9222                      & \underline{0.5093}                      & \textbf{0.2940}                     & \underline{0.4768}                      & \textbf{0.3339}                      & 0.0712                                        \\
                          & IGEV                                                       & 0.7486                   & \underline{0.6770}                   & \textbf{0.3472}                   & 1.3300                  & 2.2009                      & 0.6735                      & \underline{0.4810}                      & \textbf{0.3259}                     & \underline{0.5955}                      & \textbf{0.3432}                      & 0.0707                                        \\ \hline
\multirow{8}{*}[+0.7ex]{M2014}    & PSMNet                                                     & \underline{7.1652}                   & 12.4301                  & \textbf{6.4687}                   & 14.8883                 & 16.5317                     & 11.9619                     & \underline{11.0884}                     & \textbf{6.2982}                     & \underline{6.4352}                      & \textbf{5.9379}                      & 2.4806                                        \\
                          & GANet                                                      & \underline{5.4232}                   & 7.2175                   & \textbf{4.3756}                   & 8.8204                  & 14.5286                     & 10.6207                     & \underline{6.8133}                      & \textbf{4.3614}                     & \underline{4.3084}                      & \textbf{4.0169}                      & 1.4643                                        \\
                          & STTR                                                       & \underline{5.9919}                   & 6.2478                   & \textbf{4.8673}                   & 5.9441                  & 13.0369                     & 6.2403                      & \underline{5.8629}                      & \textbf{4.1321}                     & \underline{4.4385}                      & \textbf{4.1064}                      & 2.1278                                        \\
                          & RAFT                                                       & \underline{3.0177}                   & 5.5518                   & \textbf{2.7924}                   & 7.0148                  & 7.2556                      & 6.6081                      & \underline{6.0780}                      & \textbf{3.0382}                     & \underline{3.4659}                      & \textbf{2.5483}                      & 0.7726                                        \\
                          & LEAStereo                                                  & \underline{4.2187}                   & 5.9621                   & \textbf{3.5312}                   & 7.1150                   & 11.1104                     & \underline{6.7709}                      & 6.9652                      & \textbf{3.2800}                     & \underline{3.4835}                      & \textbf{3.1245}                      & 1.0799                                        \\
                          & ACVNet                                                     & \underline{4.5361}                   & 8.7304                   & \textbf{3.9213}                   & 11.114                  & 13.5518                     & 11.2322                     & \underline{8.6013}                      & \textbf{3.8292}                     & \underline{4.4198}                      & \textbf{3.6391}                      & 1.3322                                        \\
                          & IGEV                                                       & \underline{3.8321}                   & 5.8557                   & \textbf{3.2527}                   & 5.9323                  & 9.8204                      & 5.8438                      & \underline{5.7671}                      & \textbf{3.1207}                     & \underline{4.2789}                      & \textbf{2.8119}                      & 0.8946                                        \\ \hline
\end{tabular}%
}
\caption{The average AUC values for K2012, K2015, M2014, and VK-S6 datasets. The `Stereo' denotes the stereo-matching network. The `Optimal' denotes the AUC value of ground truth confidence map. The best and second-best results in each combination of input modalities are highlighted and underlined, respectively.}
\label{tab:quan_rslt}
\end{table*}

\begin{figure*}[!t]
    \centering
    \includegraphics[width=\linewidth]{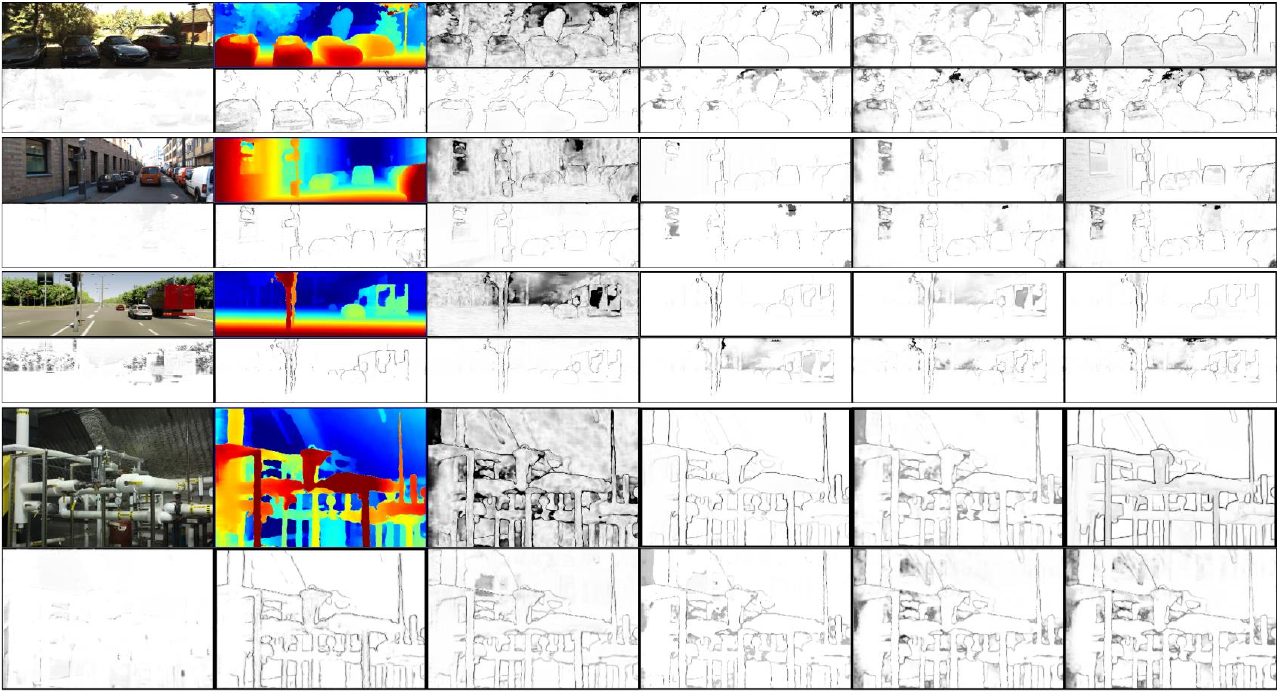}
    \caption{The confidence maps on K2012 ($1^{st}$ and $2^{nd}$ rows), K2015 ($3^{rd}$ and $4^{th}$ rows), VK2-S6 ($5^{th}$ and $6^{th}$ rows), and M2014 (last two rows) datasets using PSMNet. (From top to bottom, left to right) left image, predicted disparity map, estimated confidence maps by Ours, CCNN, CCNN$^{\dagger}$, LFN, ConfNet, LGC, LAF-Net$^*$, LAF-Net$^{*\dagger}$, LAF-Net, and LAF-Net$^{\dagger}$.}
    \label{fig:qual_rslt}
\end{figure*}

\begin{table*}[t]
\centering
\resizebox{\linewidth}{!}
{%
\begin{tabular}{clrrrrrrrrrrr}
\hline
\multicolumn{1}{c}{\multirow{2}{*}[-0.7ex]{Dataset}} & \multicolumn{1}{c}{\multirow{2}{*}[-0.7ex]{Stereo}} & \multicolumn{3}{c}{\multirow{1}{*}[-0.4ex]{Single-Modality}} & \multicolumn{5}{c}{\multirow{1}{*}[-0.4ex]{Bi-Modality}} & \multicolumn{2}{c}{\multirow{1}{*}[-0.4ex]{Tri-Modality}} & \multicolumn{1}{c}{\multirow{2}{*}[-0.7ex]{Optimal}} \\\cmidrule(rl){3-5}\cmidrule(rl){6-10}\cmidrule(rl){11-12}
                          & \multicolumn{1}{c}{} & \multicolumn{1}{c}{\multirow{1}{*}[+0.15ex]{Ours}} & \multicolumn{1}{c}{\multirow{1}{*}[+0.15ex]{CCNN}} & \multicolumn{1}{c}{\multirow{1}{*}[+0.3ex]{CCNN$^{\dagger}$}} & \multicolumn{1}{c}{\multirow{1}{*}[+0.15ex]{LFN}} & \multicolumn{1}{c}{\multirow{1}{*}[+0.15ex]{ConfNet}} & \multicolumn{1}{c}{\multirow{1}{*}[+0.15ex]{LGC-Net}} & \multicolumn{1}{c}{\multirow{1}{*}[+0.15ex]{LAF-Net$^{*}$}} & \multicolumn{1}{c}{\multirow{1}{*}[+0.3ex]{LAF-Net$^{*\dagger}$}} & \multicolumn{1}{c}{\multirow{1}{*}[+0.15ex]{LAF-Net}} & \multicolumn{1}{c}{\multirow{1}{*}[+0.3ex]{LAF-Net$^{\dagger}$}} & \multicolumn{1}{c}{}                          \\\hline
                                 
\multirow{8}{*}[+0.7ex]{\shortstack[c]{VK2-S6\\-fog}}     & PSMNet                                                     & \underline{1.5714}                     & 3.9270                   & \textbf{1.5670}                   & 4.2668                  & 4.3988                      & \underline{1.8486}                      & 2.0230                       & \textbf{1.2546}                     & \underline{1.3938}                      & \textbf{1.1294}                      & 0.3102                                        \\
                                 & GANet                                                      & \underline{0.7645}                     & 1.9908                   & \textbf{0.5007}                   & 3.0603                  & 4.1442                      & 3.1586                      & \underline{0.8559}                      & \textbf{0.3242}                     & \underline{0.6379}                      & \textbf{0.4782}                      & 0.0839                                        \\
                                 & STTR                                                       & \underline{1.1953}                     & 1.2905                   & \textbf{1.0104}                  & 3.9370                  & 3.9897                      & \underline{1.3478}                      & 1.3715                      & \textbf{1.1486}                     & \underline{0.8871}                      & \textbf{0.8606}                      & 0.2735                                        \\
                                 & RAFT                                                       & \underline{0.7572}                     & 1.7562                   & \textbf{0.5356}                   & 4.6943                  & \underline{1.1980}                      & 1.2518                      & 1.4674                      & \textbf{0.4874}                     & \underline{0.7345}                      & \textbf{0.5278}                      & 0.0918                                        \\
                                 & LEAStereo                                                  & \underline{0.5285}                     & 1.5298                   & \textbf{0.4032}                   & 2.1754                  & 2.9666                      & 3.2066                      & \underline{0.8729}                      & \textbf{0.2934}                     & \underline{0.4875}                      & \textbf{0.3315}                      & 0.0696                                        \\
                                 & ACVNet                                                     & \underline{1.0103}                     & 2.5226                   & \textbf{0.7175}                   & 7.4959                  & 4.3985                      & 5.5526                      & \underline{1.9168}                      & \textbf{0.4846}                     & \underline{0.8232}                      & \textbf{0.6108}                      & 0.1134                                        \\
                                 & IGEV                                                       & \underline{0.6386}                     & 0.8663                   & \textbf{0.2981}                   & 1.2349                  & 2.9868                      & \underline{0.6100}                      & 1.0303                      & \textbf{0.2814}                     & \underline{1.1326}                      & \textbf{0.3224}                      & 0.0657                                        \\ \hline
\multirow{8}{*}[+0.7ex]{\shortstack[c]{VK2-S6\\-morning}} & PSMNet                                                     & \underline{1.1007}                     & 1.9984                   & \textbf{0.9888}                   & 3.2150                  & 5.3708                      & 1.7921                      & \underline{1.1212}                      & \textbf{0.6655}                     & \underline{0.7872}                      & \textbf{0.7284}                      & 0.1990                                        \\
                                 & GANet                                                      & \underline{0.7917}                     & 0.8453                   & \textbf{0.5879}                   & 0.9904                  & 3.0191                      & 0.5635                      & \underline{0.5562}                      & \textbf{0.4603}                     & \underline{0.4286}                      & \textbf{0.4108}                      & 0.0677                                        \\
                                 & STTR                                                       & 0.8737                     & \underline{0.8032}                   & \textbf{0.7455}                   & 1.3725                  & 3.2641                      & \underline{0.8398}                      & 1.3731                      & \textbf{0.5812}                     & \underline{0.5825}                      & \textbf{0.5648}                      & 0.1887                                        \\
                                 & RAFT                                                       & \underline{0.5524}                     & 0.7493                   & \textbf{0.4553}                   & 1.3715                  & 1.6059                      & \underline{0.8357}                      & 1.0748                      & \textbf{0.3453}                     & \underline{0.9491}                      & \textbf{0.4893}                      & 0.0776                                        \\
                                 & LEAStereo                                                  & 0.6015                     & \underline{0.5518}                   & \textbf{0.4961}                   & 1.6568                  & 2.5703                      & 0.8983                      & \underline{0.4954}                      & \textbf{0.3562}                     & \underline{0.4910}                      & \textbf{0.3657}                      & 0.0685                                        \\
                                 & ACVNet                                                     & \underline{0.7416}                     & 1.4871                   & \textbf{0.5607}                   & 3.8478                  & 4.6123                      & 3.5231                      & \underline{1.3998}                      & \textbf{0.3796}                     & \underline{0.6356}                      & \textbf{0.4019}                      & 0.0924                                        \\
                                 & IGEV                                                       & 0.7853                     & \underline{0.7131}                   & \textbf{0.3666}                   & 1.8922                  & 3.6770                      & \underline{0.6899}                      & 1.1511                      & \textbf{0.4087}                     & \underline{0.8151}                      & \textbf{0.3590}                      & 0.0800                                        \\ \hline
\multirow{8}{*}[+0.7ex]{\shortstack[c]{VK2-S6\\-rain}}    & PSMNet                                                     & \underline{2.4996}                     & 6.2982                   & \textbf{2.3545}                   & 9.4548                  & 8.4538                      & \underline{4.5375}                      & 5.2998                      & \textbf{1.7692}                     & \underline{1.8509}                      & \textbf{1.6727}                     & 0.7111                                        \\
                                 & GANet                                                      & \underline{1.5227}                     & 2.6191                   & \textbf{1.0581}                   & 3.6266                  & 6.1154                      & 2.8830                      & \underline{2.4524}                      & \textbf{0.8130}                     & \underline{0.9638}                      & \textbf{0.9056}                      & 0.2286                                        \\
                                 & STTR                                                       & \underline{2.6439}                     & 2.8803                   & \textbf{2.2719}                   & 4.4403                  & 10.3646                     & \underline{3.1961}                      & 3.7272                      & \textbf{2.0025}                     & \underline{2.1443}                      & \textbf{1.9899}                      & 0.8483                                        \\
                                 & RAFT                                                       & \underline{2.2389}                     & 4.8650                   & \textbf{2.0608}                   & 8.6372                  & 10.7812                     & 6.8327                      & \underline{4.8524}                      & \textbf{1.8744}                     & \underline{2.7674}                      & \textbf{1.9274}                      & 0.6331                                        \\
                                 & LEAStereo                                                  & \underline{1.3635}                     & 2.1458                   & \textbf{1.1759}                   & 5.3306                  & 8.2161                      & 3.8112                      & \underline{2.5093}                      & \textbf{1.0158}                     & \underline{1.3694}                      & \textbf{1.0810}                      & 0.3362                                        \\
                                 & ACVNet                                                     & \underline{1.9070}                     & 4.9129                   & \textbf{1.5718}                   & 7.8770                  & 9.4251                      & 6.0507                      & \underline{3.7207}                      & \textbf{1.2443}                     & \underline{1.6612}                      & \textbf{1.5304}                      & 0.4879                                        \\
                                 & IGEV                                                       & \underline{1.8363}                     & 2.4547                   & \textbf{1.1263}                   & 3.9116                  & 6.8699                      & \underline{2.4990}                      & 3.6585                      & \textbf{1.1037}                     & \underline{2.5958}                      & \textbf{1.1617}                      & 0.3769                                        \\ \hline
\multirow{8}{*}[+0.7ex]{\shortstack[c]{VK2-S6\\-sunset}}  & PSMNet                                                     & \underline{1.2313}                     & 1.9764                   & \textbf{1.0755}                   & 3.4468                  & 5.2673                      & 1.8211                      & \underline{1.4248}                      & \textbf{0.7716}                     & \underline{0.8089}                      & \textbf{0.7719}                      & 0.1999                                        \\
                                 & GANet                                                      & 0.8773                     & \underline{0.8102}                   & \textbf{0.5619}                   & 0.9904                  & 3.0339                      & 0.6094                      & \underline{0.5207}                      & \textbf{0.3475}                     & \underline{0.4516}                      & \textbf{0.4204}                      & 0.0659                                        \\
                                 & STTR                                                       & 0.9846 & \underline{0.8723}                   & \textbf{0.7970}                   & 1.2677                  & 4.3465                      & \underline{0.9322}                      & 0.9880                      & \textbf{0.6915}                     & \underline{0.6521}                      & \textbf{0.6083}                      & 0.1953                                        \\
                                 & RAFT                                                       & \underline{0.5739}                     & 0.7588                   & \textbf{0.4599}                   & 0.9453                  & 1.3361                      & \underline{0.7941}                      & 0.8583                      & \textbf{0.3745}                     & \underline{1.1111}                      & \textbf{0.5814}                      & 0.0787                                        \\
                                 & LEAStereo                                                  & 0.6104                     & \underline{0.5405}                   & \textbf{0.4788}                   & 1.0449                  & 3.4167                      & 0.7635                      & \underline{0.3694}                      & \textbf{0.3185}                     & \underline{0.4908}                      & \textbf{0.3952}                      & 0.0663                                        \\
                                 & ACVNet                                                     & \underline{0.8389}                     & 1.3606                   & \textbf{0.6004}                   & 3.0129                  & 4.1981                      & 3.1625                      & \underline{0.8184}                      & \textbf{0.4192}                     & \underline{0.6201}                      & \textbf{0.4870}                      & 0.0921                                        \\
                                 & IGEV                                                       & 0.7873                     & \underline{0.7688}                   & \textbf{0.3583}                   & 1.2189                  & 3.2871                      & \underline{0.7139}                      & 0.8050                      & \textbf{0.3564}                     & \underline{0.6740}                      & \textbf{0.3628}                      & 0.0816                                        \\ \hline
\end{tabular}%
}
\caption{The average AUC values for VK2-S6 dataset with 4 different weather conditions (fog, morning, rain, and sunset). The `Stereo' denotes the stereo-matching network. The `Optimal' denotes the AUC value of ground truth confidence map. The best and second-best results in each combination of input modalities are highlighted and underlined, respectively.}
\label{tab:quan_rslt_vk2}
\end{table*}

\subsection{Datasets}
\textbf{KITTI 2012 (K2012)}~\cite{KITTI2012} and \textbf{KITTI 2015 (K2015)}~\cite{KITTI2015}, which are outdoor driving scene datasets, consist of 194 and 200 stereo-image pairs and corresponding sparse ground truth disparity maps obtained from LiDAR sensor measurements, respectively.
We split the K2012 into 20 images for training and 174 images for testing following~\cite{poggi_conf_setting}.
\textbf{Virtual KITTI 2 (VK2)}~\cite{VKITTI2}, which is a photo-realistic virtual driving scene dataset, contains 21,260 stereo-image pairs of different 6 scenes with various weather and illumination conditions (fog, overcast, rain, morning, and sunset).
We only use Scene06 (VK2-S6) for testing, which is referred to test dataset by the authors.
\textbf{Middlebury 2014 (M2014)}~\cite{Middlebury} is an indoor scene dataset, which is composed of high-resolution 15 stereo pair images and corresponding dense ground truth disparity maps.
For the M2014, we use quarter-resolution images in all experiments following~\citet{poggi_conf_setting}.
We train all stereo-confidence networks using the K2012 training set (20 images) and evaluate them on the K2012 test set (174 images), K2015 (200 images), VK2-S6 (270 images), and M2014 (15 images).
Also, we experiment with challenging subsets of VK2-S6 with four different weather and illumination conditions (fog, rain, morning, and sunset).
We exclude pixels with disparities $d>192$ in training and test in all datasets.

\subsection{Stereo-Matching Networks}
To demonstrate the adaptability, end-to-end stereo-matching networks with various architectures and state-of-the-art performance, such as PSMNet~\cite{PSMNet}, GANet~\cite{GANet}, STTR~\cite{STTR}, LEAStereo~\cite{LEAStereo}, RAFT-Stereo~\cite{RAFT-Stereo}, ACVNet~\cite{ACVNet}, and IGEV-Stereo~\cite{IGEV-Stereo} are used to obtain predicted disparity maps and cost volume, which are used as input modalities of stereo-confidence estimation networks.
Regardless of the dataset and network type, we use weights fine-tuned on the KITTI datasets provided by authors for all experiments.

\subsection{Confidence Networks}
As classified in Table~\ref{tab:input_mod}, we set CCNN~\cite{CCNN}, LFN~\cite{LFN}, ConfNet~\cite{LGC-Net}, LGC-Net~\cite{LGC-Net}, LAF-Net$^*$~\cite{LAF-Net}, and LAF-Net~\cite{LAF-Net} as comparison groups of the proposed method considering various combinations of input modalities. 
These methods still show state-of-the-art performance as mentioned in~\cite{ConfidenceSurvey,meta-confidence}.
The LAF-Net$^*$ denotes LAF-Net that uses bi-modal inputs (disparity, image) without cost volume.
For CCNN$^{\dagger}$, LAF-Net$^{*\dagger}$, and LAF-Net$^{\dagger}$, which use the proposed method (Ours) as an additional input modality, the modifications are limited to fundamental aspects such as concatenating input modalities or adding a few layers.
Among the existing stereo-confidence methods using bi-modal inputs, since LAF-Net$^{*}$ generally shows better performance than LFN, ConfNet, and LGC-Net, we experiment with LAF-Net$^{*}$ to check the usefulness of Ours as an additional input modality (LAF-Net$^{*\dagger}$).
We obtain all experimental results using codes provided by authors without modifying any hyperparameters.
All experiments are conducted on a machine with 8 GeForce RTX 2080 Ti GPUs.

\subsection{Evaluation Metrics} 
As introduced in \citet{stereo_conf}, we evaluate the performance of each stereo-confidence method by an area under the curve (AUC) value of the ROC curve, which represents how well the measurement identifies correct matches.
Ideally, if the measurement identifies all correct matches, the optimal AUC value can be obtained as 
\begin{equation}\int_{1-\varepsilon}^{1}{\frac{x-\left (1-\varepsilon\right)}{x}dx=\varepsilon+\left (1-\varepsilon \right )\textup{ln}\left(1-\varepsilon \right )},\end{equation}
where $\varepsilon$ denotes the error rate computed over the entire disparity map.
As in~\citet{poggi2021confidence}, we set the threshold value $\tau$ to 3 in obtaining ground truth confidence maps.
The optimal AUC values are measured by using ground truth confidence maps.
All reported AUC values in tables are multiplied by a factor of $10^{2}$ to ease the readability.

\begin{table}[t!]
\centering
\resizebox{\columnwidth}{!}
{
\begin{tabular}{rrrrr}
\hline
\multicolumn{1}{c}{\multirow{2}{*}[-0.15ex]{Method}} & \multicolumn{1}{c}{\multirow{2}{*}[-0.15ex]{\# Params.}} & \multicolumn{1}{c}{\multirow{2}{*}[-0.15ex]{AUC $\downarrow$}} & \multicolumn{2}{c}{\multirow{1}{*}[-0.2ex]{Latency (s)}}                        \\ \cline{4-5} 
\multicolumn{1}{c}{}                        & \multicolumn{1}{c}{}                                & \multicolumn{1}{c}{}                          & \multicolumn{1}{c}{\multirow{1}{*}[-0.3ex]{1 GPU}} & \multicolumn{1}{c}{\multirow{1}{*}[-0.3ex]{$N$ GPUs}} \\ \hline
Ours ($N$=2, $k_0$=0, $k_1$=1)                                        & \multicolumn{1}{c}{-}                                                   & 0.7917                                        & 0.6490                    & 0.6429                     \\
Ours ($N$=3, $K$=1)                                        & \multicolumn{1}{c}{-}                                                   & 0.5455                                        & 0.9301                    & 0.7435                     \\
Ours ($N$=5, $K$=2)                                        & \multicolumn{1}{c}{-}                                                   & \underline{0.4729}                                        & 1.5114                    & 0.8551                     \\
Ours ($N$=7, $K$=3)                                        & \multicolumn{1}{c}{-}                                                   & \textbf{0.4383}                                        & 2.0735                    & 0.9959                     \\ \hline
LAF-Net$^*$                                     & 0.57M                                                & \underline{0.7820}                                        & 0.6552                    & \multicolumn{1}{c}{-}                          \\
LAF-Net$^{*\dagger}$ ($N$=5)                                     & 0.69M                                                & \textbf{0.3563}                                        & 1.6446                    & 0.9590                     \\ \hline
LAF-Net                                     & 0.69M                                                & \underline{0.3862}                                        & 0.7325                    & \multicolumn{1}{c}{-}                          \\
LAF-Net$^{\dagger}$ ($N$=5)                                     & 0.80M                                                & \textbf{0.3414}                                        & 1.7014                    & 1.1915                     \\ \hline
\end{tabular}}

\caption{Further study for the average AUC value and the latency according to the number of disparity plane shifts $N$ on K2012 using RAFT. The best and second-best results in each experiment are highlighted and underlined, respectively.}
\label{tab:abl_N}
\end{table}

\begin{table}[t!]
\centering
\resizebox{0.8\columnwidth}{!}{%
\begin{tabular}{ccccc}
\hline
\multirow{1}{*}[-0.25ex]{Method}     & \multirow{1}{*}[-0.25ex]{$K$=1}                        & \multirow{1}{*}[-0.25ex]{$K$=2}                        & \multirow{1}{*}[-0.25ex]{$K$=4}                        & \multirow{1}{*}[-0.25ex]{$K$=8}                        \\ \hline
\multirow{1}{*}[-0.25ex]{Ours ($N$=3)} & \multicolumn{1}{r}{\multirow{1}{*}[-0.25ex]{\textbf{0.5455}}} & \multicolumn{1}{r}{\multirow{1}{*}[-0.2ex]{\underline{0.6022}}} & \multicolumn{1}{r}{\multirow{1}{*}[-0.25ex]{0.6158}} & \multicolumn{1}{r}{\multirow{1}{*}[-0.25ex]{0.6984}} \\ \hline
\end{tabular}%
}
\caption{Ablation study for the shifting step size on K2012 using RAFT. The best and second-best results are highlighted and underlined, respectively.}
\label{tab:abl_K}
\end{table}

\subsection{Confidence Estimation Analysis}
\textbf{Quantitative results.}
In Table~\ref{tab:quan_rslt}, we compare the average AUC values of Ours and existing learning-based methods on the K2012, K2015, VK2-S6, and M2014 datasets to validate the effectiveness of Ours.
Although Ours is conventional and uses a single-modality input, it shows competitive results compared to existing methods in most cases regardless of the datasets and stereo-matching networks.

Although internal features such as cost volume are not feasible to be used in the stereo-confidence networks in safety-critical systems, there are some exceptions where the cost volume can be utilized.
For such cases, we experiment with LAF-Net using Ours as an additional input modality. 
Without exceptions, when Ours is used as an additional input, LAF-Net$^\dagger$ shows better performance than LAF-Net.

To further check compatibility with the existing learning-based stereo-confidence methods, we also experiment using Ours as an additional input modality to CCNN and LAF-Net$^{*}$.
CCNN$^{\dagger}$ and LAF-Net$^{*\dagger}$ also show better performance than CCNN and LAF-Net$^{*}$.
It demonstrates that Ours can be utilized as a useful input even for learning-based stereo-confidence networks. 
Surprisingly, there are various cases in which LAF-Net$^{*\dagger}$ shows better performance than LAF-Net even if the cost volume is not utilized.
It indicates that Ours properly reinterprets the conventional stereo-confidence methods to be suitable for end-to-end stereo-matching networks by analyzing the disparity profile out of the network.
We believe Ours can serve as an alternative to the cost volume as an input modality in the existing and future learning-based stereo-confidence networks.

\noindent\textbf{Qualitative results.}
In Fig.~\ref{fig:qual_rslt}, we visualize the qualitative results of Ours and the existing methods on the K2012, K2015, VK2-S6, and M2014.
Although the existing methods generally show strong capabilities in detecting occlusion boundaries, they struggle to identify other remaining ill-posed regions like repeated patterns, textureless regions, and non-Lambertian surfaces.
Ours effectively identifies not only inaccurate but also ill-posed regions such as occlusion boundaries, repeated patterns, textureless regions, and non-Lambertian surfaces.
See supplementary material for more results.

\subsection{Further Studies}
\textbf{Different weather conditions.}
In Table~\ref{tab:quan_rslt_vk2}, we also evaluate Ours and existing methods on VK2-S6 of four different weather conditions (fog, morning, rain, sunset) for further studies on various driving environments.
The performances of VK-S6-fog and VK-S6-rain are generally worse than those of VK-S6-morning and VK-S6-sunset because VK-S6-fog and VK-S6-rain cannot be observed in the K2012 dataset, which is a training dataset.
Similar to Table~\ref{tab:quan_rslt}, the methods with the proposed method, which are denoted to superscript $\dagger$, generally show better performance than the other methods, even in different weather conditions, especially in VK-S6-fog and VK-S6-rain.
See supplementary material for more qualitative results.

\noindent\textbf{The trade-off between AUC value and latency.}
In Table~\ref{tab:abl_N}, we examine the influence of varying the number of disparity plane shifts $N$ on the average AUC value and latency using the input samples from the K2012 dataset with full resolution ($H\times W=384\times 1248$) and the RAFT stereo-matching network.
Since the latency of Ours depends on $N$, the latency shown in Table~\ref{tab:abl_N} is composed of the processing time of RAFT and that of the stereo-confidence methods.

Ours requires a minimum of two disparity maps to work. 
With the minimum requirement setting $N=2$, it exhibits marginally reduced performance but boasts faster latency on both single and multiple GPUs compared to LAF-Net$^*$. 
As $N$ increases, the average AUC value of Ours consistently improves, but at the cost of increased latency. 
By adjusting the batch size to $N$ during inference, $N$ disparity maps can be obtained by a single forward from RAFT. 
A clear limitation of Ours is the slower latency under resource-constrained environments, like using a single GPU. 
Yet, this can be mitigated by scaling the number of GPUs in tandem with $N$. 
To break it down: LAF-Net$^*$ has a latency comprised of 0.5817 seconds for RAFT processing and 0.0735 seconds for LAF-Net$^*$ itself. 
In contrast, Ours ($N=2$) with a single GPU consists of 0.6490 seconds for RAFT and a negligible time from $10^{-5}$ to $10^{-6}$ seconds for our own processing time.
The latency is averaged over K2012 test set (174 images).

\noindent\textbf{Varing shifting step size $K$.}
As shown in Table~\ref{tab:abl_K}, we further examine the performance according to the shifting step size $K$. 
With $N=3$, the performance is typically decreased as the shifting step size $K$ is increased. 
Thus, we fix the shifting step size $K$ at the minimum pixel unit of 1.

\begin{table}[t!]
\centering
\resizebox{\columnwidth}{!}
{
\begin{tabular}{cccccc}
\hline
\multirow{1}{*}[-0.25ex]{ConfNet}                    & \multirow{1}{*}[-0.25ex]{OTB}                        & \multirow{1}{*}[-0.25ex]{OTB-online}                 & \multirow{1}{*}[-0.1ex]{ConfNet$^{\dagger}$}        & \multirow{1}{*}[-0.25ex]{Ours}                        & \multirow{1}{*}[-0.25ex]{Optimal}                    \\ \hline
\multicolumn{1}{r}{\multirow{1}{*}[-0.25ex]{1.7420}} & \multicolumn{1}{r}{\multirow{1}{*}[-0.25ex]{1.7419}} & \multicolumn{1}{r}{\multirow{1}{*}[-0.25ex]{1.4632}} & \multicolumn{1}{r}{\multirow{1}{*}[-0.25ex]{\textbf{0.9581}}} & \multicolumn{1}{r}{\multirow{1}{*}[-0.2ex]{\underline{1.3013}}} & \multicolumn{1}{r}{\multirow{1}{*}[-0.25ex]{0.0654}} \\ \hline
\end{tabular}}
\caption{Comparison to self-adaptation method on DrivingStereo using GANet. The `Optimal' denotes the AUC value of ground truth confidence map. The best and second-best results are highlighted and underlined, respectively.}
\label{tab:comp_self}
\end{table}

\noindent\textbf{Comparison to self-adaptation method}
Self-adapting confidence~\cite{selfadapting} is close to Ours.
While the self-adapting confidence is rooted in a conventional approach, it is built upon the learning-based framework without using internal information from stereo-matching methods.
The self-adapting confidence is achieved by integrating three stereo-confidence cues: image reprojection error, agreement among neighboring matches, and a uniqueness constraint.
Table~\ref{tab:comp_self} presents a comparison between Ours and self-adapt confidence, which denoted OTB and OTB-online. 
The experiments utilize GANet on 6905 samples of the DrivingStereo dataset \cite{DrivingStereo} provided by authors, as following the procedures laid out in the self-adapt confidence method \cite{selfadapting}.
Both OTB and OTB-online are trained in a self-supervised manner and OTB-online is OTB with online adaptation.
Since OTB and OTB-online are implemented on ConfNet, the performance of ConfNet is also presented as a baseline for the supervised method. 
While both OTB and OTB-online outperform ConfNet, our method surpasses the results of both.
ConfNet$^\dagger$ stands out with the best performance, but it is trained under supervision, making direct comparisons inequitable.

\section{Conclusion}
We present a stereo-confidence measurement that operates outside end-to-end stereo-matching networks, which is a recent paradigm of the learning-based stereo-matching methods. 
The key idea of the proposed method is to reinterpret the conventional stereo-confidence method analyzing the cost profile to be suitable for end-to-end stereo-matching networks by analyzing disparity profiles. 
To measure the confidence, using a predicted disparity map without the disparity plane sweep as an anchor, the desirable disparity profile shaped into a linear line is generated and compared to the disparity maps obtained using the disparity plane sweep.
We also investigate compatibility with the learning-based stereo-confidence networks using the proposed method as an additional input modality.
Our extensive experimental results demonstrate that the proposed method not only shows competitive confidence performance but also significantly enhances the performance of learning-based methods as an additional input modality.
In future research, we plan to utilize the proposed method in the disparity refinement framework working in a self-supervised manner.

\section{Acknowledgements}
This work was supported by Institute of Information \& communications Technology Planning \& Evaluation (IITP) grant funded by the Korea government(MSIT) [NO.2022-0-00184, Development and Study of AI Technologies to Inexpensively Conform to Evolving Policy on Ethics]

\bibliography{aaai24}

\end{document}